\documentclass[11pt,a4paper]{article}
\usepackage[hyperref]{naaclhlt2018}
\usepackage{times}
\usepackage{latexsym}

\usepackage{booktabs} 

\newcommand{\0}{\hspace*{0.5em}}

\usepackage{graphicx}
\usepackage{verbatim}
\usepackage{enumerate}

\usepackage{todonotes}

\usepackage{url}

\aclfinalcopy 
\def\aclpaperid{6} 

\title{A Pipeline for Creative Visual Storytelling}

\author{Stephanie M. Lukin, Reginald Hobbs, Clare R. Voss \\
  U.S. Army Research Laboratory \\
  Adelphi, MD, USA \\
  {\tt stephanie.m.lukin.civ@mail.mil } }  

\date{}

\begin{document}
\maketitle
\begin{abstract}
Computational visual storytelling produces a textual description of events and interpretations depicted in a sequence of images. These texts are made possible by advances and cross-disciplinary approaches in natural language processing, generation, and computer vision. We define a computational creative visual storytelling as one with the ability to alter the telling of a story along three aspects: to speak about different environments, to produce variations based on narrative goals, and to adapt the narrative to the audience. These aspects of creative storytelling and their effect on the narrative have yet to be explored in visual storytelling. This paper presents a pipeline of task-modules, Object Identification, Single-Image Inferencing, and Multi-Image Narration, that serve as a preliminary design for building a creative visual storyteller. We have piloted this design for a sequence of images in an annotation task. We present and analyze the collected corpus and describe plans towards automation.

\end{abstract}

\section{Introduction}

\begin{figure*}[h!]
    \centering
        \includegraphics[width=0.9\textwidth]{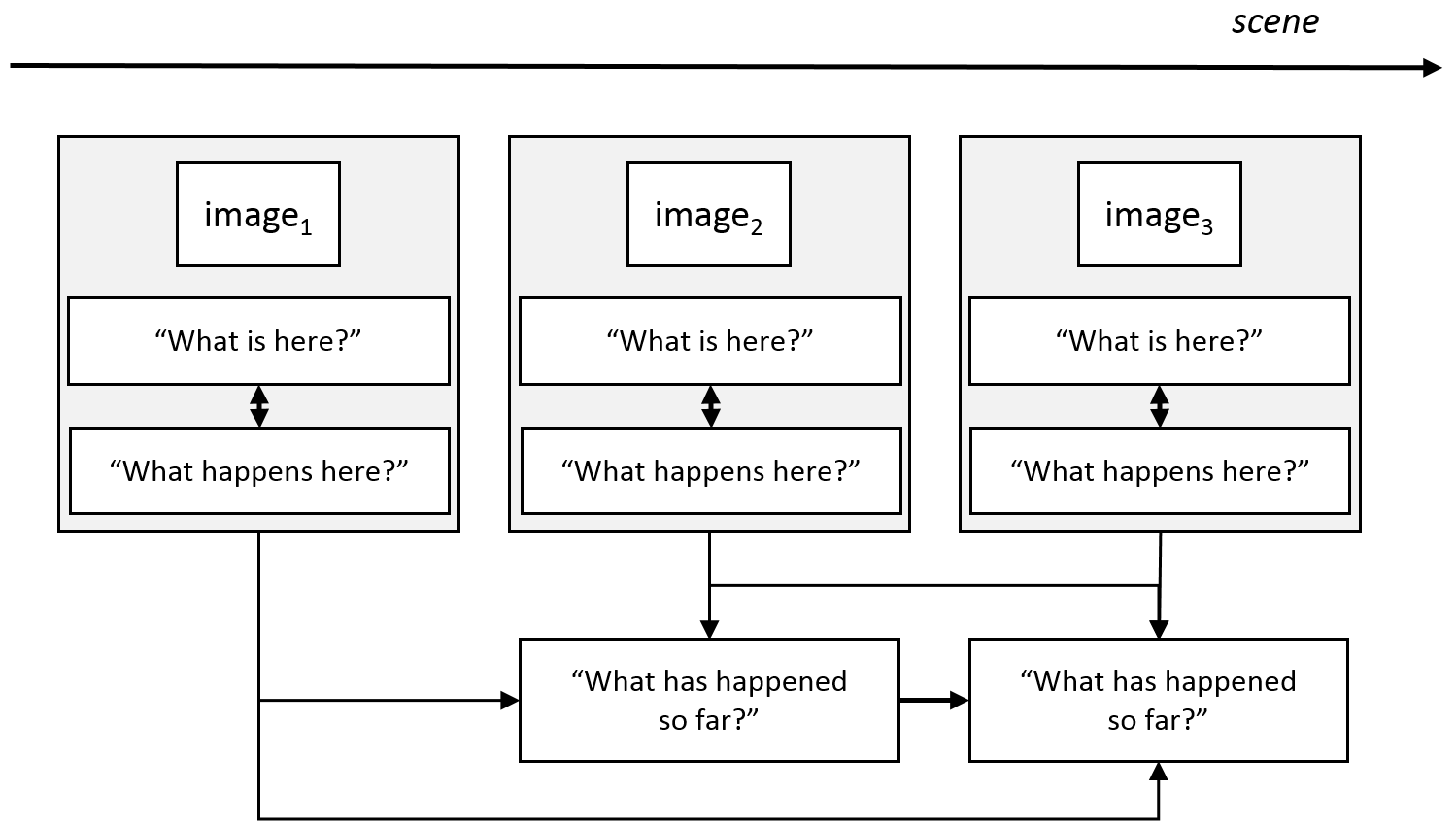}
    \caption{Creative Visual Storytelling Pipeline: T1 (Object Identification), T2 (Single Image Inferencing), \\ T3 (Multi-Image Narration)}\label{fig:pipeline}
    \vspace{-0.1in}
\end{figure*}

Telling stories from multiple images is a creative challenge that involves visually analyzing the images, drawing connections between them, and producing language to convey the message of the narrative. To computationally model this creative phenomena, a visual storyteller must take into consideration several aspects that will influence the narrative: the environment and presentation of imagery \cite{madden200699}, the narrative goals which affect the desired response of the reader or listener \cite{bohanek2006family,thorne2003telling}, and the audience, who may prefer to read or hear different narrative styles \cite{thorne1987press}.

The environment is the content of the imagery, but also its interpretability (e.g., image quality). Canonical images are available from a number of high-quality datasets \cite{everingham2010pascal,plummer2015flickr30k,
lin2014microsoft,ordonez2011im2text}, however, there is little coverage of low-resourced domains with low-quality images or  atypical camera perspectives that might appear in a sequence of pictures taken from blind persons, a child learning to use a camera, or a robot surveying a site. For this work, we studied an environment with odd surroundings taken from a camera mounted on a ground robot.

Narrative goals guide the selection of what objects or inferences in the image are relevant or uncharacteristic. The result is a narrative tailored to different goals such as a general ``describe the scene'', or a more focused ``look for suspicious activity''. The most salient narrative may shift as new information, in the form of images, is presented, offering different possible interpretations of the scene.  
This work posed a forensic task with the narrative goal to describe what may have occurred within a scene, assuming some temporal consistency across images. This open-endedness evoked creativity in the resulting narratives. 

The telling of the narrative will also differ based upon the target audience. A concise narrative is more appropriate if the audience is expecting to hear news or information, while a verbose and humorous narrative is suited for entertainment. Audiences may differ in how they would best experience the narrative: immersed in the first person or through an omniscient narrator. The audience in this work was unspecified, thus the audience was the same as the storyteller defining the narrative. 

To build a computational creative visual storyteller that customizes a narrative along these three aspects, we propose a creative visual storytelling pipeline requiring separate task-modules for Object Identification, Single-Image Inferencing, and Multi-Image Narration. We have conducted an exploratory pilot experiment following this pipeline to collect data from each task-module to train the computational storyteller. The collected data provides instances of creative storytelling from which we have analyzed what people see and pay attention to, what they interpret, and how they weave together a story across a series of images.

Creative visual storytelling requires an understanding of the creative processes.
We argue that existing systems cannot achieve these creative aspects of visual storytelling. Current object identification algorithms may perform poorly on low-resourced environments with minimal training data. Computer vision algorithms may over-identify objects, that is, describe more objects than are ultimately needed for the goal of a coherent narrative. Algorithms that generate captions of an image often produce generic language, rather than language tailored to a specific audience. Our pilot experiment is an attempt to reveal the creative processes involved when humans perform this task, and then to computationally model the phenomena from the observed data. 

Our pipeline is introduced in Section~\ref{sec:pipeline}, where we also discuss computational considerations and the application of this pipeline to our pilot experiment. In Section~\ref{sec:pilot} we describe the exploratory pilot experiment, in which we presented images of a low-quality and atypical environment and have annotators answer ``what may have happened here?" This open-ended narrative goal has the potential to elicit diverse and creative narratives. We did not specify the audience, leaving the annotator free to write in a style that appeals to them. The data and analysis of the pilot are presented in Section~\ref{sec:results}, as well as observations for extending to crowdsourcing a larger corpus and how to use these creative insights to build computational models that follow this pipeline. In Section~\ref{sec:related} we compare our approach to recent works in other storytelling methodologies, then conclude and describe future directions of this work in Section~\ref{sec:conc}.

\section{Creative Visual Storytelling Pipeline}
\label{sec:pipeline}

The pipeline and interaction of task-modules we have designed to perform creative visual storytelling over multiple images are depicted in Figure~\ref{fig:pipeline}. Each task-module answers a question critical to creative visual storytelling: ``what is here?'' (T1: Object Identification), ``what happens here?'' (T2: Single-Image Inferencing), and ``what has happened so far?'' (T3: Multi-Image Narration). We discuss the purpose, expected inputs and outputs of each module, and explore computational implementations of the pipeline.

\subsection{Pipeline}
This section describes the task-modules we designed that provide answers to our questions for creative visual storytelling.

{\bf Task-Module 1: Object Identification (T1).} Objects in an image are the building blocks for storytelling that answer the question, literally, ``what is here?'' This question is asked of every image in a sequence for the purposes of object curation. From a single image, the expected outputs are objects and their descriptors. We anticipate that two categories of object descriptors will be informative for interfacing with the subsequent task-modules: spatial descriptors, consisting of object {\it co-locations} and {\it orientation}, and observational {\it attribute} descriptors, including color, shape, or texture of the object. Confidence level will provide information about the {\it expectedness} of the object and its descriptors, or if the object is difficult or {\it uncertain} to decipher given the environment. 


{\bf Task-Module 2: Single-Image Inferencing (T2).} Dependent upon T1, the Single-Image Inferencing task-module is a literal interpretation derived from the objects previously identified in the context of the current image. 
After the curation of objects in T1, a second round of content selection commences in the form of inference determination and selection. Using the selected objects, descriptors, and expectations about the objects, this task-module answers the question ``what happens here?'' 
For example, the function of ``kitchen'' might be extrapolated from the co-location of a cereal box, pan, and crockpot.

Separating T2 from T1 creates a modular system where each task-module can make the best decision given the information available. However, these task-modules are also interdependent: as the inferences in T2 depend upon T1 for object selection, so too does the object selection depend upon the inferences drawn so far.

\begin{figure*}[h!]
    \centering
        \includegraphics[width=0.32\textwidth]{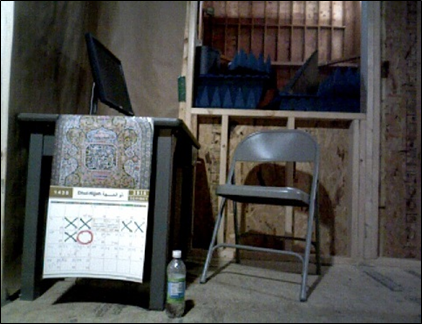}
        \includegraphics[width=0.33\textwidth]{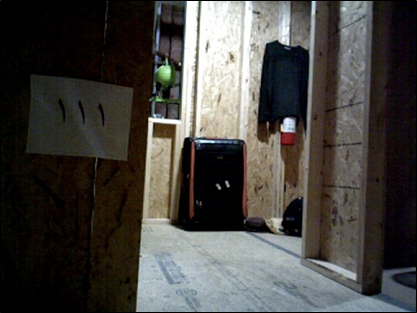}
        \includegraphics[width=0.33\textwidth]{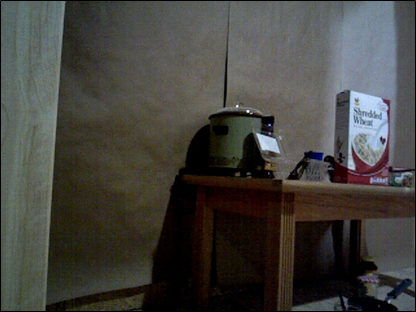}
    \caption{image$_1$, image$_2$, and image$_3$ in Pilot Experiment Scene}\label{fig:b507}
\end{figure*}


{\bf Task-Module 3: Multi-Image Narration (T3).}
A narrative can indeed be constructed from a single image, however, we designed our pipeline to consider when additional context, in the form of additional images, is provided. The Multi-Image Narration task-module draws from T1 and T2 to construct the larger narrative. All images, objects, and inferences are taken into consideration when determining ``what has happened so far?'' and ``what has happened from one image to the next?'' 
This task-module performs narrative planning by referencing the inferences and objects from the previous images. It then produces a natural language output in the form of a narrative text. Plausible narrative interpretations are formed from global knowledge about how the addition of new images confirm or disprove prior hypotheses and expectations.

\subsection{From Pipeline Design to Pilot}

Our first step towards building this automated pipeline is to pilot it. We will use the dataset collected and the results from the exploratory study to to build an informed computational, creative visual storyteller. When piloting, we refer to this pipeline a sequence of annotation tasks.

T1 is based on computer vision technology. Of particular interest are our collected annotations on the low-quality and atypical environments that traditionally do not have readily available object annotations. Commonsense reasoning and knowledge bases drive the technology behind deriving T2 inferences. T3 narratives consist of two sub-task-modules: narrative planning and natural language generation. Each technology can be matched to our pipeline, and be built up separately, leveraging existing works, but tuned to this task.

Our annotators are required to write in natural language (though we do not specify full sentences) the answers to the questions posed in each task-module. While this natural language intermediate representation of T1 and T2 is appropriate for a pilot study, a semantic representation of these task-modules might be more feasible for computation until the final rendering of the narrative text. For example, drawing inferences in T2 with the objects identified in T1 might be better achieved with an ontological representation of objects and attributes, such as WordNet \citep{fellbaum1998wordnet}, and inferences mined from a knowledge base. 

In our annotation, the sub-task-modules of narrative planning and natural language generation are implicitly intertwined. The annotator does not note in the exercise intermediary narrative planning before writing the final text. In computation, T3 may generate the final narrative text word-by-word (combining narrative planning and natural language generation). Another approach might first perform narrative planning, followed by generation from a semantic or syntactic representation that is compatible with intermediate representations from T1 and T2.

\section{Pilot Experiment}
\label{sec:pilot}

A paper-based pilot experiment implementing this pipeline was conducted. Ten annotators (A$_1$~-~A$_{10}$)\footnote{A$_5$, an author of this paper, designed the experiment and examples. All annotators had varying degrees of familiarity with the environment in the images.} participated in the annotation of the three images in Figure~\ref{fig:b507} (image$_1$~-~image$_3$). These images were taken from a camera mounted on a ground robot while it navigated an unfamiliar environment. 
The environment was static, thus, presenting these images in temporal order was not as critical as it would have been if the images were still-frames taken from a video or if the images contained a progression of actions or events. 

Annotators first addressed the questions posed in the Object Identification (T1) and Single-Image Inference (T2) task-modules for image$_1$. They repeated the process for image$_2$ and image$_3$, and authored a Multi-Image Narrative (T3). The annotator work flow mimicked the pipeline presented in Figure~\ref{fig:pipeline}. For each subsequent image, the time allotted increased from five, to eight, to eleven minutes to allow more time for the narrative to be constructed after annotators processed the additional images. An example image sequence with answers was provided prior to the experiment. A$_5$ gave a brief, oral, open-ended explanation of the experiment as not to bias annotators to what they should focus on in the scene or what kind of language they should use.
The goal of this data collection is to gather data that models the creative storytelling processes, not to track these processes in real-time. A future web-based interface will allow us to track the timing of annotation, what information is added when, and how each task-module influences the other task-modules for each image. 

Object Identification did not require annotators to define a bounding box for labeled objects, nor were annotators required to provide objective descriptors\footnote{As we design a web-based version of this experiment, we will enforce interfaces explicitly linked to object annotations, and the desire to view previously annotated images.}.  Annotators authored natural language labels, phrases, or sentences to describe objects, attributes, and spatial relations while indicating confidence levels if appropriate.  

During Single Image Inferencing, annotators were shown their response from T1 as they authored a natural language description of activity or functions of the image, as well as a natural language explanation of inferences for that determination, citing supporting evidence from T1 output. For a single image, annotators may answer the questions posed by T1 and T2 in any order to build the most informed narrative. 

Annotators authored a Multi-Image Narrative to explain what has happened in the sequence of images presented so far. For each image seen in the sequence, annotators were shown their own natural language responses from T1 and T2 for those images.  Annotators were encouraged to look back to their responses in previous images (as the bottom row of Figure~\ref{fig:pipeline} indicates), but not to make changes to their responses about the previous images. They were, however, encouraged to incorporate previous feedback into the context of the current image. From this task-module, annotators wrote a natural language narrative connecting activity or functions in the images which will be used to learn how to weave together a story across the images.

The open-ended ``what has happened here?'' narrative goal has no single answer. These annotations may be treated as ground truth, but we run the risk of potentially missing out on creative alternatives. Bootstraping all possible objects and inferences would achieve greater coverage, yet this quickly becomes infeasible. We lean toward the middle, where the answers collected will help determine what annotators deem as important.

\begin{table}[t]
\centering
\small
\begin{tabular}{cp{1.5in}}
\toprule
\# Annotators & Objects \\ \midrule
10 & calendar, water bottle \\
9 & computer, table/desk \\ 
8 & chair \\
4 & walls, window \\
2 & blue triangles \\
1 & floor, praying rug \\
\bottomrule
\end{tabular}
\caption{\label{obj1counts}Objects identified by annotators in image$_1$}
\vspace{0.2in}

\begin{tabular}{cp{1.5in}}
\toprule
\# Annotators & Objects \\ \midrule
10 & suitcase, shirt \\
8 & sign \\ 
6 & green object \\
5 & fire extinguisher \\
4 & walls \\
3 & bag \\
2 & floor, window \\
1 & coat hanger, shoes, rug \\ 
\bottomrule
\end{tabular}
\caption{\label{obj2counts}Objects identified by annotators in image$_2$}
\vspace{0.2in}

\begin{tabular}{cp{1.5in}}
\toprule
\# Annotators & Objects \\ \midrule
7 & crockpot, cereal box, table \\
6 & pan \\
5 & container \\
2 & walls, label \\
1 & thread and needle, coffee pot, jam, door frame \\
\bottomrule
\end{tabular}
\caption{\label{obj3counts}Objects identified by annotators in image$_3$ (total of 7 annotators)}
\label{obj3}
\end{table}

\section{Results and Analysis}
\label{sec:results}

In this section, we discuss and analyze the collected data and provide insights for incorporating each task-module into a computational system.

\subsection{Object Identification (T1)}

Thirty three objects were identified across the images.\footnote{Due to time constrains, A$_2$~-~A$_4$ did not complete image$_3$.} A$_5$ identified the most of these objects (20), and A$_1$, the least (10). Tables~\ref{obj1counts}~-~\ref{obj3counts} show the objects identified and how many annotators referenced each object. A set of objects emerged in each image that captured the annotators' attention. Object descriptor categories are tabulated in Table~\ref{tab:objects}\footnote{Tabulation of descriptors in Tables~\ref{obj1}~-~\ref{obj3} in Appendix.}. Not surprisingly, the most common descriptors were attributes, e.g., color and shape, followed by co-locations. Orientation was not observed in this dataset, however this category may be useful for other disrupted environments. We observed instances of uncertainty, e.g., ``a suitcase, not entirely sure, because of zipper and size'', and unexpected objects, ``unfinished floor'', whereas ``floors'' may have not been labeled otherwise. 

Lack of coverage and overlap in this task with respect to objects and descriptors is not discouraging. In fact, we argue that exhaustive object identification is counter-intuitive and detrimental to creative visual storytelling. Annotators may have  identified only the objects of interest to the narrative they were forming, and viewed other objects as distractors. The most frequent of the identified objects are likely to be the most influential in T2 where the calendar, computer, and chair provide more information than the ``blue triangles''. 

Not only can selective object identification provide the most salient objects for deriving interpretations, but the Object Identification exercise with respect to storytelling can differentiate between objects and descriptors that are commonplace or otherwise irrelevant. For instance, if a fire extinguisher was not annotated as red, we are inclined to deduce it is because this fact is well known or unimportant, rather than the result of a distracted annotator.\footnote{We expect this to be revealed in the web-based version of the task with a stricter annotation interface.}

When automating this task-module, new object identification algorithms should account for the following: a sampling of relevant objects specific to the storytelling challenge, and attention to potential outlier descriptors which may be more indicative than a standard descriptor, depending on the environment.

\begin{table}[t]
	\centering
	\begin{small}
		\begin{tabular}{p{0.76in}p{0.21in}lll}
			\toprule
			& Total & Average & Min & Max  \\ \midrule
			{\bf Spatial} & & & & \\
			\hspace*{1em}Co-Location & 51 & 6.3 & 0 & 14   \\ \midrule
			{\bf Observational} & & & & \\
			\hspace*{1em}Attribute & 99 & 12.2 & 3 & 22  \\ \midrule
			{\bf Confidence} & & & & \\
			\hspace*{1em}Unexpected & 7 & 0.7 & 0 &  4 \\ 
			\hspace*{1em}Uncertainty & 28 & 3.3 & 0 & 8 \\ \midrule
			{\bf Total} & 185 & 22.6 & 7 & 39 \\ \bottomrule
		\end{tabular}
		\caption{Object descriptor summary with counts per annotator (A$_2$~-~A$_4$ excluded from average, min, and max; see footnote 4)\label{tab:objects}}
	\end{small}
\end{table}

\begin{table*}[t!]
\centering
\begin{small}
\begin{tabular}{|l|p{2.6in}|p{2.6in}|}
\hline
Image & Single-Image Inference & Multi-Image Narrative \\ \hline \hline
Image$_1$ 
&
Looks like a dingy, sparse office. The {\it computer desk}, {\it calendar }indicate an office, but the space is unfinished ({\it no dry wall, carpet}) and area outside {\it window} looks weird, not like an office building. 
& \\ 
\hline \hline
Image$_2$ 
& 
Looks like someone was staying here temporarily, using this now to store {\it clothes}, or maybe as a bedroom. Again, it's atypical because its an {\it unfinished space} that looks uncomfortable. 
& 
I think this person was hiding out here to get ready for some event. The space isn't finished enough to be intended for habitation, but someone had to stay here, perhaps because they didn't want to be found, and you wouldn't expect someone to be living in a construction zone. \\ 
\hline \hline
Image$_3$
&
This area was used as a sort of kitchen or {\it food storage} prep area. 
&
Someone was definitely living here even though it wasn't finished or intended to be a house. They were probably using a crock pot because you can make food in this without having larger appliances like a stove, oven. There's no milk, so this person may be lactose intolerant. The robot should vanquish them with milk. \\ 
\hline
\end{tabular}
\caption{\label{a1}A$_1$'s annotation (previously identified objects in Single-Image Inference text in italics)}
\vspace{0.2in}

\begin{tabular}{|l|p{2.6in}|p{2.6in}|}
\hline
Image & Single-Image Inference & Multi-Image Narrative \\ \hline \hline
Image$_1$ 
&
This is likely a workplace of some sort. It is unclear if it is an {\it unfinished part} of a current/suspended construction project or it is just a utilitarian space inside of an industrial facility. The presence of a {\it computer monitor} suggest it is in use or a low crime area. 
& \\ 
\hline \hline
Image$_2$
&
This is a jobsite of some sort. It has {\it unfinished walls} and what may be a {\it paper shredder}. 
&
This is an unfinished building. There is some evidence of office-type work (i.e. work involving paper and computers). The existence of ``windows'' between rooms suggests that this is not a dwelling (or intended to become one), that is, a building designed to be a dwelling, but what it is remains unclear. \\ 
\hline \hline
Image$_3$
&
A room in a building is being used as a cooking and eating station, based upon presence of {\it food, table}, and {\it cooking instruments}. 		
&
This building is being used by a likely small number of individuals for unclear purposes including cooking, eating, and basic office work. \\ \hline
\end{tabular}
\caption{\label{a8}A$_8$'s annotation (previously identified objects in Single-Image Inference text in italics)}
\end{small}
\end{table*}

\subsection{Single-Images Inferencing (T2)}

We highlight A$_1$ and A$_8$ for the remainder of the discussion\footnote{Other annotation results in Tables~\ref{a2}~-~\ref{a10} in Appendix.}. Table~\ref{a1} shows A$_1$'s annotation of Single-Image Inferencing and Multi-Image Narration. In the Single-Image Inferencing (T2) for image$_1$, A$_1$ noted the ``office'' theme by referencing the desk and computer, and expressed uncertainty with respect to the window looking ``weird'' and unlike a typical office building. A$_1$ kept clear the distinction between images in their annotation of image$_2$, as there were no references to the office observed only in image$_1$. Instead, references in image$_2$ were to the storage of clothes.
In the single-image interpretation of image$_3$, A$_1$ suggested that this was a food preparation area from the presence of the crockpot, cereal, and the other food items that appeared together. 
A$_8$, whose annotation is in Table~\ref{a8}, also noted the ``workplace'' theme from the desk and computer, though A$_8$ leaned more towards a construction site, citing the utilitarian space. Due to uncertainty of the environment, A$_8$ misidentified the suitcase in image$_2$ as a shredder, and incorporated it prominently into their interpretation. Similar to A$_1$, A$_8$ also indicated in image$_3$ that this was a food preparation area.

A$_8$'s misinterpretation of the suitcase raises an implementation question: are the inferences and algorithms we develop only as good as our environment data allows them to be? How might a misunderstanding of the environment affect the inferences? This environment showcased the uniqueness of the physical space and low-quality of images, yet all annotators indicated, without prompting or instruction, varying degrees of confidence in their interpretations based upon the evidence. A$_8$ indicated their uncertainty about the suitcase object by hedging that it was ``what may be a paper shredder''. This expression of uncertainty should be preserved in an automated system for instances such as this when an answer is unknown or has a low confidence level. 

T2 is intended to inform a commonsense reasoner and knowledge base based on T1 to deduce the setting. This task-module describes functions of rooms or spaces, e.g., food preparation areas and office space. Additional interpretations about the space were made by annotators from the overall appearance of objects in the image, such as the atmospheric observation ``lighting of rooms is not very good'' (A$_7$, Table~\ref{a7} in Appendix). These inferences might not be easily deducible from T1 alone, but the combination of these task-modules allows for these to occur.

Evaluating this annotation in a computational system will require some ground truth, though we have previously stated that it is impossible to claim such a gold standard in a creative storytelling task. Evaluation must therefore be subject to both qualitative and quantitative analyses, including, but not limited to, commonsense reasoning on validation sets and determining plausible alternatives to commonsense interpretations.

\subsection{Multi-Image Narration (T3)}

The narrative begins to form across the first two images in the Multi-Image Narration task-module (T3). A$_1$ hypothesized that someone was ``hiding out'', going a step beyond their T2 inference of an ``office space'' in image$_1$, to extrapolate ``what has happened here'' rather than ``what happens here''. In image$_2$, A$_1$ had hedged their narrative with ``I think'', but the language became stronger and more confident in image$_3$, in which A$_1$ ``definitely'' thought that the space was inhabited. A$_1$ pointed out that a lack of milk was unexpected in a canonical kitchen, and supplemented their narrative with a joke, suggesting to ``vanquish them with milk''.
In image$_2$, A$_8$ interpreted that the space was not intended for long-term dwelling. Their narrative shifted in image$_3$ when another scene was revealed. A$_8$ concluded that this space was inhabited by a group, despite the annotator's previous assumption in image$_2$ that it was not suited for this purpose.

There is no a guaranteed ``correct'' narrative that unfolds, especially if we are seeking creativity. Some narrative pieces may fall into place as additional images provided context, but in the case of these environments, annotators were challenged to make sense of the sequence and pull together a plausible, if not uncertain, narrative.

The narrative goal and audience aspects of creative visual storytelling will directly inform T3. A variety of creative narratives and interpretations emerged from this pilot, despite the particularly sparse and odd environment and openness of the narrative goal. Based on the responses from each successive task-modules, all annotators' interpretations and narratives are correct. Even with annotator misunderstandings, the narratives presented were their own interpretation of the environment. 
As the audience in this task was not specified, annotators could use any style to tell their story. The data collected expressed creativity through jokes (A$_1$), lists and structured information (A$_5$), concise deductions (A$_6$, A$_8$), uncertain deductions (A$_4$), first person (A$_1$, A$_3$, A$_5$), omniscient narrators (A$_2$), and the use of ``we'' inclusive of the robot navigating the space (A$_7$, A$_9$, A$_{10}$). 

Future annotations may assign an audience or a style prompt in order to observe the varied language use. This will inform computational models by curating stylistic features and learning from appropriate data sources.

\section{Related work}
\label{sec:related}

Visual storytelling is still a relatively new subfield of research
that has not yet begun to capture the highly creative stories 
generated by text-based storytelling systems to date. 
The latter supports the definition of specific goals or 
presents alternate narrative interpretations by generating stories according to character goals 
(e.g., \citet{meehan1977tale}) and author goals (e.g., \citet{lebowitz1985story}). Other interactive, co-constructed, text-based narrative systems make use of information retrieval methods by implicitly linking the text generation to the interpretation. As a result, systems incorporating these methods cannot be adjusted for different narrative goals or audiences  \cite{cychosz2017effective,swanson2008say,munishkina2013fully}. 

Other research in text-based storytelling focuses on answering the question ``what happens next?'' to infer the selection of the most appropriate next sentence. This method indirectly relies on the selection of sentences to evaluation the results of a forced choice between the ``best'' or ``correct'' next sentence of the choices when given a narrative context 
(as in the Story Close Test \citep{mostafazadeh2016corpus} 
and the Children's Book Test \citep{hill2015goldilocks}). 
Our pipeline, by contrast, builds on a series of open-ended questions, 
for which there is no single gold-standard or reference answer. 
Instead, we expect in time to follow prior work by \citet{roemmele2011choice}
where evaluation will entail generating and ranking plausible interpretations.

Recent work on caption generation combines computer vision with a simplified narration, or single sentence text description of an image \citep{vinyals2015show}. Image processing typically takes place in one phase, while text generation follows in a second phase. Superficially, this separation of phases resembles the division of labor in our approach, where T1 and T2 involve image-specific analysis, and T3 involves text generation. 
However this form of caption generation depends solely on training data where individual images are paired with individual sentences.
It assumes the T3 sub-task-modules can be learned from the same data source, and generates the same sentences on a per-image basis, regardless of the order of images. One can readily imagine the inadequacy of stringing together captions to construct a narrative, where the same captions describe both images of a waterfall flowing down, and those same images in reverse order where instead the water seems to be flowing up.

The work most similar in approach to our visual storyteller annotation pipeline is \citet{huang2016visual} who separate their tasks into three tiers: 
the first over single images, generating literal descriptions of images in isolation (DII), the second over multiple images, generating literal descriptions of images in sequence (DIS), and the third over multiple images, generating stories for images in sequence (SIS). While these tiers may seem analogous to ours, there are different assumptions underlying the tasks in data collection. 
For each task, their images are annotated independently by different annotators, 
while in our approach, all images are annotated by annotators performing all of our tasks. 
The DII task is an exhaustive object identification task on single images, yet we leave T1 up to our annotators to determine how many objects and attributes to describe in an image to avoid the potential for object over-identification.
The SIS task involves a set of images over which annotators select and possibly reorder, then write one sentence per image to create a narrative, with the opportunity to skip images. In our pipeline, we have intentionally designed our task-modules to allow for the possibility of one task-module to build off of and influence one another. It is possible in our approach for an annotator's inference in T2 of one image to feed forward and affect their T1 annotations in the subsequent image, which might in turn affect the resulting T3 narrative. 
In short, \citet{huang2016visual} capture the thread of storytelling in one tier only, their SIS condition, while our annotators build their narratives across task-modules as they progress from image to image.

\section{Conclusion and Future Work}
\label{sec:conc}

This paper introduces a creative visual storytelling pipeline for a sequence of images that delegates separate task-modules for Object Identification, Single-Image Inferencing, and Multi-Image Narration. These task-modules can be implemented to computationally describe diverse environments and customize the telling based on narrative goals and different audiences. The pilot annotation has collected data for this visual storyteller in a low-resourced environment, and analyzed how creative visual storytelling is performed in this pipeline for the purposes of training a computational, creative visual storyteller. The pipeline is grounded in narrative decision-making processes, and we expect it to perform well on both low- and high-quality datasets. Using only curated datasets, however, runs the risk of training algorithms that are not general use.

We are now positioned to conduct a crowdsourcing annotation effort, followed by an implementation of this storyteller following the outlined task-modules for automation. Our pipeline and implementation detail are algorithmically agnostic. We anticipate off-the-shelf and state-of-the-art computer vision and language generation methodologies will provide a number of baselines for creative visual storytelling: 
to test environments, compare an object identification algorithm trained on high-quality data against one trained on low-quality data;
to test narrative goals, compare a computer vision algorithm that may over-identify objects against one focused on a specific set to form a story; 
to test audience, compare a caption generation algorithm that may generate generic language against one tailored to the audience desires. 

The streamlined approach of our experimental annotation pipeline allows us to easily prompt for different narrative goals and audiences in future crowdsourcing to obtain and compare different narratives. Evaluation of the final narrative must take into consideration the narrative goal and audience. In addition, evaluation must balance the correctness of the interpretation with expressing creativity, as well as the grammaticality of the generated story, suggesting new quantitative and qualitative metrics must be developed.  


\bibliography{storynlp2018}
\bibliographystyle{acl_natbib} 

\newpage

\appendix
\section*{Appendix: Additional Annotations}
\vspace{-0.2in}
\begin{table}[h!]
\centering
\small
\begin{tabular}{|p{0.9in}|l|p{3.8in}||l|l|}
\hline
Object & \# & Descriptor Text & \# & Descriptor \\ \hline \hline
Calendar & 10
	& hanging off the table, 
	taped to table top & 4 & co-location \\
	& & marked up, 
	ink, red circle on calendar, marked with pen & 4 & attribute \\
	& & foreign language & 1 & attribute \\	
	& & paper & 1 & attribute \\ 
	& & picture on top & 1 & attribute \\ \hline
Water bottle & 10
	& on the floor, on ground, on floor & 3 & co-location \\
	& & to the right of table & 1 & co-location \\
	& & mostly empty, unclear if it has been opened & 2 & attribute \\ 
	& & plastic & 2 & attribute \\
	& & closed with lid & 1 & attribute \\ \hline
Computer & 9
	& screen, black turned off; monitor, black & 3 & attribute \\ \hline
Table / Desk & 9
	& has computer on it, [computer] on table & 2 & co-location \\ 
	& & gray, black & 2 & attribute \\
	& & wood & 2 & attribute \\
	& & metal & 1 & attribute \\
	& & (presumed) rectangular & 1 & uncertainty \\ \hline
Chair & 8
	& folding & 6 & attribute \\ 
	& & metal & 5 & attribute \\ 
	& & grey & 3 & attribute \\ \hline
Walls & 4 
	& wood & 1 & attribute \\
	& & unfinished and showing beams, unfinished construction & 3 & unexpected \\ \hline
Window & 4
	& in wall behind chair & 1 & co-location \\  \hline
	& & window to another room; perhaps chairs in other room & 1 & uncertainty \\
	& & no glass & 1 & unexpected \\ \hline
Blue triangles & 2
	& blue objects in windowsill & 1 & unexpected \\ \hline
Floor & 1 
	& unfinished & 1 & unexpected \\ \hline
Praying rug & 1 
	& & & \\ \hline
\end{tabular}
\vspace{-0.1in}
\begin{minipage}{\textwidth}
\label{obj1}
\caption{Object Identification for image$_1$}
\end{minipage}
\vspace{0.2in}

\begin{tabular}{|p{0.9in}|l|p{3.8in}||l|l|}
\hline
Object & \# & Descriptor Text & \# & Descriptor \\ \hline \hline
Suitcase & 10
	& black, orange stripes; black and red; black with red trim; blue and copper & 5 & attribute \\
	& & (not entirely sure) because of zipper item and size & 1 & uncertainty \\
	& & resembles a paper shredder & 1 & uncertainty \\ 
	& & a suitcase or a heater & 1 & uncertainty \\ \hline
Shirt & 10
	& on hanger; on fire extinguisher; on wall; hanging & 6 & co-location \\
	& & black & 5 & attribute \\
	& & long sleeves & 2 & attribute \\
	& & black thing hanging on wall (unclear what it is); black object & 2 & uncertainty \\ \hline
Sign & 8
	& on the wall & 4 & co-location \\
	& & maybe indicating '3'?; roman numerals; 3 dashes; Arabic numbers; foreign language; room number 111 & 6 & attribute \\
	& & poster & 1 & attribute \\
	& & map or blueprints & 1 & uncertainty \\ \hline
Green object & 6 
	& spherical & 1 & attribute \\
	& & hanging in window; in windowsill & 2 & co-location \\
	& & green thing outside room; green object; unidentifiable object; lime green object & 4 & uncertainty \\
	& & light post? fan? & 1 & uncertainty \\ \hline
Fire extinguisher & 5
	& hanging off of black thing (also unclear as to what this is or does); on wall & 3 & co-location \\
	& & obscured & 1 & co-location \\ 
	& & cylindrical & 1 & attribute \\ 
	& & white and red thing; red object, white and red piece of object & 3 & uncertainty \\ \hline
Wall & 4
	& wooden & 1 & attribute \\ 
	& & unfinished; visible plywood studs & 2 & unexpected \\ \hline
Bag & 3 
	& backpack or bag; something round; pile of clothes & 2 & uncertainty \\ 
	& & on the ground & 2 & co-location \\ \hline
	& & next to suitcase & 1 & co-location \\ \hline
Floor & 2 
	& marking of industry grade particle board, unfinished & 2 & attribute \\ \hline
Window & 2 
	& & & \\ \hline
Coat hanger & 1  
	& hanging on wall & 1 & co-location \\
	& & wire & 1 & attribute \\ 
	& & white & 1 & attribute \\ \hline
Shoes & 1  
	& shoes or hat & 1 & uncertainty \\ \hline
Rug & 1  
	& & & \\ \hline
\end{tabular}
\vspace{-0.5in}
\begin{minipage}{\textwidth}
\caption{Object Identification for image$_2$}
\end{minipage}
\label{obj2}
\end{table}

\newpage


\begin{table*}
\small
\begin{tabular}{|p{0.9in}|l|p{3.8in}||l|l|}
\hline
Object & \# & Descriptor Text & \# & Descriptor \\ \hline \hline
Crockpot & 7\0
	& on table & 1 & co-location \\ 
	& & green & 2 & attribute \\ 
	& & old fashioned & 1 & attribute \\ 
	& & kitchen appliance & 1 & attribute \\ 
	& & white or silver & 1 & uncertainty \\\hline
Cereal box & 7\0
	& on table & 1 & co-location \\ 
	& & to the right of crockpot & 1 & co-location \\
	& & shredded wheat & 4 & attribute \\
	& & cardboard & 1 & attribute \\
	& & printed black letters & 1 & attribute \\ \hline
Table & 7\0
	& wood & 3 & attribute \\
	& & coffee table style & 2 & attribute \\ 
	& & pale & 1 & attribute \\ \hline
Pan & 6\0
	& on ground; on floor & 3 & co-location \\ 
	& & blue handle & 1 & attribute \\ 
	& & medium-size & 1 & attribute \\ \hline
Container & 5\0
	& clear & 2 & attribute \\
	& & plastic & 3 & attribute \\ 
	& & empty & 2 & attribute \\
	& & rectangular & 1 & attribute \\ 
	& & hinged top & 1 & attribute \\ \hline
Walls & 2\0
	& lined with paper & 1 & attribute \\ \hline
Label & 2\0
	& on pressure cooker & 2 & co-location \\
	& & white & 1 & attribute \\ \hline
Thread and needle & 1\0
	& to the right of cereal box & 1 & co-location  \\ \hline
Coffee pot & 1\0
	& what looks like a coffee pot & 1 & uncertainty \\
	& & empty & 1 & attribute \\
	& & behind cereal box & 1 & attribute \\ \hline
Jam & 1\0
	& plaid red and white lid & 1 & attribute  \\ \hline
Door frame & 1\0
	& & &  \\ \hline
\end{tabular}
\caption{Object Identification for image$_3$}
\vspace{0.1in}
\label{obj3}

\centering
\begin{tabular}{|l|p{2.6in}|p{2.6in}|}
\hline
Image & Single-Image Inferencing & Multi-Image Narration \\ \hline \hline
Image$_1$ 
&
Someone sits at table and puts water bottle on floor while perhaps taking notes for others in some room. Folding chair suggests temporary or new use of space while building under construction.
& \\ 
\hline \hline
Image$_2$
& 
Hallway view, suggesting exit path where someone might leave luggage while being in building 
& 
Same building as in the first scene because same type of wood for walls, floor, and opening/window construction. Arabic numbers on paper sign loosely attached (because wavy surface of paper e.g. not rigid, not laminated) to the wall suggests temporary designation of space for specific use, as an organized arrangement by some people for others. \\ 
\hline \hline
Image$_3$
&
N/A
&
N/A \\ 
\hline
\end{tabular}
\caption{\label{a2}A$_2$'s annotation}
\vspace{0.1in}

\begin{tabular}{|l|p{2.6in}|p{2.6in}|}
\hline
Image & Single-Image Inferencing & Multi-Image Narration \\ \hline \hline
Image$_1$ 
&
I believe this is an office, because there is a computer monitor on a table, the table is serving as a desk, and there is a metal chair next to the monitor and the desk. A calendar is typically found in an office, however the calendar here is not in a location that is convenient for a person & \\\hline \hline
Image$_2$
&
I believe that this is a standard room that serves as a storage area. The absence of other objects does not hint at this room serving any other purpose. 
&
I believe that this storage room is located in a home since personal items such as a luggage bag and spare shirt are not typically found in a public building. From the marked calendar in the previous picture, it appears that the occupants are preparing to travel very soon. \\ 
\hline \hline
Image$_3$
&
N/A
&
N/A\\ \hline
\end{tabular}
\caption{\label{a3}A$_3$'s annotation}
\vspace{0.1in}
\end{table*}


\begin{table*}[t!]
\centering
\begin{small}
\begin{tabular}{|l|p{2.6in}|p{2.6in}|}
\hline
Image & Single-Image Inferencing & Multi-Image Narration \\ \hline \hline
Image$_1$ 
&
An office or computer setting/workstation. Actual computer maybe under desk (not visible) or missing. Water bottle suggests someone used this space recently. Chair not facing desk suggests person left in a hurry (not pushed under desk). Red circled date suggests some significance. & \\ \hline
\hline 
Image$_2$
&
Shirt and suitcase suggests someone stored their personal items in this space. Room being labeled suggests recent occupants used more than 1 part of this space. Space does not look comfortable, but personal effects are here anyway. Holiday? 
&
Someone camped out here and planned activities. They left in a hurry and didn't spend time putting things in their suitcases, or they had a visitor and the visitor left abruptly. The occupant may have left on the date marked in the calendar. The date may have had personal significance for an operation. \\ 
\hline \hline
Image$_3$
&
N/A	
&
N/A \\ \hline
\end{tabular}
\caption{\label{a4}A$_4$'s annotation}
\vspace{0.1in}

\centering

\begin{tabular}{|l|p{2.6in}|p{2.6in}|}
\hline
Image & Single-Image Inferencing & Multi-Image Narration \\ \hline \hline
Image$_1$ 
&
Office (chair, desk, computer, calendar). Unfinished building (walls, floor, window) & \\\hline \hline
Image$_2$
&
In an unfinished building closet, common space. Things thrown to the side. Doesn't care much about office safety because fire extinguisher is covered, therefore not easily accessible. 
&
Not sure about either workzone because randomly placed clothes and unsafe work environment. Could be a factory with unsafe conditions. Someone living or storing clothes in a ``break room''? \\ 
\hline \hline
Image$_3$
&
``Camp'' site but not outdoors. Items on floor indicate some disarray or disregard for cleanliness. Why is the crock pot on the coffee table with cereal? Breakfast? But why are the walls strange?
&
Food like this shouldn't appear in a safe work environment, so I no longer think that. Someone seems to be living here in an unsafe and probably unregulated (re: fire extinguisher) way. Someone is hiding out in an uninhabited warehouse or work site (walls, floors, windows) \\ \hline
\end{tabular}
\caption{\label{a5}A$_5$'s annotation}
\vspace{0.1in}

\centering
\begin{tabular}{|l|p{2.6in}|p{2.6in}|}
\hline
Image & Single-Image Inferencing & Multi-Image Narration \\ \hline \hline
Image$_1$ 
&
This is an office space because there is a desk, chair, computer and calendar. These items are typical items that would be in an office space. & \\\hline \hline
Image$_2$
&
This looks like a storage space, a closet, or the entrance/exit to a building. People typically pile things such as a suitcase, hanging clothes, backpack, etc. at one of those locations. A storage space or closet would allow for the items to be stored for a long time but would also be due to people being ready to leave on travel.
&
Due to the lack of decorations I would say these pictures were taken in a location where people were staying or working temporarily (like a headquarters safe house, etc.) \\ 
\hline \hline
Image$_3$
&
These are items that would typically be found in a kitchen or break area. You would see a table or counter in a kitchen or break room. The pan and crock pot are not items that would be seen in other rooms, like a living room, office, bathroom, bedroom.
&
I would say this is a house or temporary space because the items are not organized and the surrounding area is not decorative. The scenes look messy and it doesn't look like it gets cleaned or has been cleaned recently. Plus the space contains a suitcase which gives the impressions that the person has not unpacked. \\ \hline
\end{tabular}
\caption{\label{a6}A$_6$'s annotation}
\vspace{0.1in}
\end{small}
\end{table*}


\begin{table*}[t!]
\begin{small}
\centering
\begin{tabular}{|l|p{2.6in}|p{2.6in}|}
\hline
Image & Single-Image Inferencing & Multi-Image Narration \\ \hline \hline
Image$_1$ 
&
Looks like a place to work, with chair, table, monitor. Calendar is out of place because people don't have calendars from the edge of a table, so it can only be seen from the floor. Walls are unfurnished, only wood and plywood. A window in the wall, like an interior window. Not sure what it is a window for-why is the window in that location?  & \\ \hline
\hline 
Image$_2$
&
We are looking through a doorway or hallway. Shirt and suitcase belong together. Not sure what other objects are (green, red, black on ground).
&
Might be same location as Image 1, because the wooden/plywood walls and floor are similar. Not sure what the images have to do with each other, but might be 2 different rooms in same location. We're viewing this image from another room, because this room has a poster in it. Lighting of rooms is not very good, almost looks like spot lights, so not like an ordinary, prototypical house. \\ 
\hline \hline
Image$_3$
&
A bunch of objects on a table, with a few objects underneath. The objects on/under the table all have to do with food or preparing food. Walls are light colored. In the foreground appears to be a wooden door jam. Although there are some kitchen items, this does not look like a typical kitchen 
&
It is difficult to tell if this is in the same location as the previous 2 images. The wood door jam might be the same, but hard to know if wall is plywood and we don't see any other wooden framing. Rooms from all 3 images don't appear connected physically. No understandable context or connections. \\ \hline
\end{tabular}
\caption{\label{a7}A$_7$'s annotation}
\vspace{0.1in}

\centering
\begin{tabular}{|l|p{2.6in}|p{2.6in}|}
\hline
Image & Single-Image Inferencing & Multi-Image Narration \\ \hline \hline
Image$_1$ 
&
 This looks like a make-shift room or space. Has a military of intel feel to it. Could be a briefing or an interrogation room. Given the prayer rug, definitely interaction between parties of different backgrounds, etc.  & \\ \hline \hline
Image$_2$
&
This view or room reflects living quarters. Given the nature of the condition of the wall, it is a make-shift. The existence of a number identifying this room indicates that it is one of many.
&
Combining the 2 pictures, this is beginning to look like part of a structure used for military/intel purposes. The location is most likely somewhere in the Middle East given how the numbers are written in Hindi indicating Arabic language. This also means we have multi-party/individual interactions. \\ 
\hline \hline
Image$_3$
&
This picture has all the ingredients to presenting a kitchen: food and cookware leads to a kitchen. Given the ``rough'' look of the setting, this has the hallmarks of a make-shift kitchen.
&
This confirms, more than anything else, the scenario described in picture 2. As a whole, looks like some sort of post or output or a make-shift temporary type. Only necessities are present and the place couldn't quickly be abandoned. \\ \hline
\end{tabular}
\caption{\label{a9}A$_9$'s annotation}
\vspace{0.1in}

\begin{tabular}{|l|p{2.6in}|p{2.6in}|}
\hline
Image & Single-Image Inferencing & Multi-Image Narration \\ \hline \hline
Image$_1$ 
& 
This was probably used as a workspace, given the chair and table with the monitor and the calendar. Someone was recently there because the bottle is upright. 
 & \\ \hline
\hline 
Image$_2$
&
This was a space that someone lived in given the clothes, fan(?), heater/suitcase(?). Given the mess, they left abruptly. The fire extinguisher indicates a presence because it is a safety aid.
&
This suggests we're in a space occupied by someone because of the office type and ``living room'' type room setup. It was purposefully made and left very abruptly (messy clothes, chair not pushed in). \\ 
\hline \hline
Image$_3$
&
This seems to be a kitchen area because all objects are food related. It is messy. The rice cooker has a blue light and may be on. There is a window letting in light, visible on the back wall.
&
This supports the assumption that the environment was recently occupied. Food is opened, rice cooker is on, mess suggests it was abruptly abandoned, much like image 2's mess. The robot appears to be I the doorway at an angle. \\ \hline
\end{tabular}
\caption{\label{a10}A$_{10}$'s annotation}
\vspace{0.1in}
\end{small}
\end{table*}

\end{document}